\documentclass[conference]{IEEEtran}
\IEEEoverridecommandlockouts
\usepackage{cite}
\usepackage{amsmath,amssymb,amsfonts}
\usepackage{algorithmic}
\usepackage{graphicx}
\usepackage{textcomp}
\usepackage{xcolor}
\usepackage{multirow}
\def\BibTeX{{\rm B\kern-.05em{\sc i\kern-.025em b}\kern-.08em
    T\kern-.1667em\lower.7ex\hbox{E}\kern-.125emX}}

\author{\IEEEauthorblockN{Jonathan C. Balloch, Julia Kim, Mark O. Riedl}
\IEEEauthorblockA{\textit{College of Computing} \\
\textit{Georgia Institute of Technology}\\
\{balloch, julia.kim, riedl\}@gatech.edu}
\and
\IEEEauthorblockN{Jessica L. Inman}
\IEEEauthorblockA{\textit{Georgia Tech Research Institute} \\
jessica.inman@gtri.gatech.edu}
}

\usepackage{hyperref}       
\usepackage{amsfonts}       
\usepackage{xcolor}         
\usepackage{colortbl}         

\title{The Role of Exploration for Task Transfer in Reinforcement Learning}

\begin{document}


\maketitle

\begin{abstract}
The exploration–exploitation trade-off in reinforcement learning (RL) is a well-known and much-studied problem that balances greedy action selection with novel experience, and the study of exploration methods is usually only considered in the context of learning the optimal policy for a single learning task.
However, in the context of online task transfer, where there is a change to the task during online operation, we hypothesize that exploration strategies that anticipate the need to adapt to future tasks can have a pronounced impact on the efficiency of transfer.
As such, we re-examine the exploration–exploitation trade-off in the context of transfer learning. 
In this work, we review reinforcement learning exploration methods, define a taxonomy with which to organize them, analyze these methods' differences in the context of task transfer, and suggest avenues for future investigation. 
\end{abstract}

\begin{IEEEkeywords}
Reinforcement Learning, Transfer Learning, Exploration, World Models
\end{IEEEkeywords}

\section{Introduction}

It has been observed in many types of animals and humans that prior knowledge helps them to learn new tasks more efficiently and effectively~\cite{perkins1992transfer, hopfield2010neurodynamics}. 
This adaptability of knowledge for decision making is vital to surviving and thriving when our world or our needs change. 
Deep reinforcement learning (RL) has driven robot learning huge leaps forward, including in generalization and adaptability. 
However, the ability of an RL agent trained for a specific task to rapidly transfer prior knowledge to a new domain or task is still inefficient or intractable, often times resulting in catastrophic forgetting~\cite{mccloskey1989catastrophic}. 
%

Transfer learning is built on the assumption that knowledge learned in an initial task can generalize to a new task. 
This is powerful because it dramatically reduces the dependence on large amounts of task-specific data to produce usable performance and makes training new models much more efficient than training \textit{tabula rasa}, or without prior knowledge~\cite{agarwal2022beyond}. 
Transfer learning is commonly studied in supervised and semi-supervised learning settings.  
In reinforcement learning, transfer is complicated by several factors:
\begin{enumerate}
\item The transfer of prior knowledge in the context of a Markov decision process, instead of a stationary data domain as in supervised learning, means that observations from different RL domains are not identically and independently distributed (i.i.d); as such, it is even less likely that observations in different reinforcement learning tasks will be similarly distributed. 
\item Because the feedback signal from reward in RL is often sparse compared to supervised or semi-supervised learning, the already-challenging credit assignment problem becomes harder when transferring between tasks, as even similar, transferable knowledge may be differently rewarded in different domains. 
\item Lastly, because the transition dynamics and reward functions of RL tasks are usually not known in advance and must be sampled as a through a process, it is difficult if not impossible to determine the similarity of two different tasks.
\end{enumerate}

Within the exploration-exploitation trade-off in single task reinforcement learning, exploration is seen as a ``necessary evil''; exploration beyond what is strictly needed for exploitation is typically assumed to be wasted computation. This assumption, however, no longer holds when RL is not optimizing for just a single task.
We hypothesize that the method of exploration in the source domain plays an important role in how well an agent transfers its policy and knowledge to the target domain.
To support transfer learning in the real world, we consider the case where the optimal policy for the source task is not the same as the optimal policy for the target task.
Indeed, the action trajectory required to perform the target task may not overlap with any sequence of actions required to perform the source task. 
Regardless, the agent may have encountered---and learned through exploration---aspects of the source domain that, despite not being useful to the source task, are in fact useful to exploring and learning in the target domain.  

In this paper, we re-examine RL's exploration-exploitation trade-off in the context of online task transfer in reinforcement learning, and question what exploration approaches should be further researched when considering transfer efficiency and performance in reinforcement learning .
We define a taxonomy with which to organize and assess the potential effectiveness of broad types of exploration methods, argue 
how the different aspects of the taxonomy make methods more or less suited for online task transfer in reinforcement learning, and suggest how this taxonomy might be used to drive future research into exploration for open world task transfer. 

\section{Background}

Reinforcement learning typically models an environment as a Markov decision process (MDP): $M=\langle \mathcal{S}, \mathcal{A}, R, P, \gamma\rangle$, 
where $\mathcal{S}$ is the space of environment states, $\mathcal{A}$ is the space of actions, $R: \mathcal{S} \times \mathcal{A} \rightarrow \mathbb{R}$ is the function that maps states and actions to a scalar reward, $P: \mathcal{S} \times \mathcal{A} \rightarrow \mathcal{S}$ is the transition function between states, and $\gamma$ is the discount factor of future reward. The learning task in RL is to model a policy $\pi(s)$ that, given any state, can select the action that maximizes either the likelihood of getting a reward or the value of an action given future reward. For brevity, we show this learning process with Q-learning, where the policy of best action is determined by the maximum value of the state-action value function $Q(s, a)$. The optimal solution $Q_*(s, a)$ is defined through the Bellman Optimality Equation:

\begin{equation}
    Q_*(s, a)=R(s, a)+\gamma \sum_{s^{\prime} \in \mathcal{S}} P_{s s^{\prime}}^a \max _{a^{\prime} \in \mathcal{A}} Q_*\left(s^{\prime}, a^{\prime}\right)
\end{equation}

\noindent where $s^{\prime}$ and $a^{\prime}$ are the next state and action, respectively~\cite{sutton2018reinforcement}. There are many ways to derive an error function to solve this objective, for example by TD-learning where the update is a learning rate weighted mixture of the estimated next-state return $R_{t+1}+\gamma \max _{a^{\prime} \in \mathcal{A}} Q\left(s_{t-1}, a\right)$ and the estimated value $Q\left(s_t, a_t \right)$. In deep reinforcement learning, a neural network of parameters $\theta$ approximates the value function, $Q_* \approx Q(S, A ; \theta)$. The parameters of this network are updated by calculating the loss function: 

\begin{equation}
\mathcal{L}(\theta)=\mathbb{E}_{\left(s, a, r, s^{\prime}\right) \sim U(B)}\left[\left(r+\gamma \max _{a^{\prime}} \hat{Q}\left(s^{\prime}, a^{\prime}\right)-Q(s, a ; \theta)\right)^2\right]
\end{equation}

\noindent where $U(B)$ is a uniform distribution for sampling random transition tuples from an experience replay buffer $B$, and $\hat{Q}(\cdot)$ is a frozen copy of the Q-function that acts as surrogate target Q-function (this target network is updated less frequently to reduce correlations between the target and current Q-functions~\cite{mnih2015human}). Even with this process, exploration is necessary to keep the policy out of local minima; by mixing the greedy selection of maximum-value actions and exploring the environment, one can be more confident that the policy is not missing the most optimal path to the goal~\cite{sutton2018reinforcement}.    

The study of transfer learning in RL, while not as widespread as in other machine learning subfields, has seen growing body of work over the past two decades~\cite{taylor2009transfer,Lazaric2012,zhu2020transfer}. 
For example, one application of great interest to the RL community is as a solution to the sim-to-real problem~\cite{chen2021cross}; however most of these works presume some knowledge about (or have access to) the target domain in advance. 


Given the similarity of model implementation and architecture between deep supervised learning and deep reinforcement learning, transfer for deep reinforcement learning is in principle equally attainable. 
A challenge unique to transfer in reinforcement learning is that \textit{similarity} between tasks is very difficult to measure and model~\cite{Lazaric2012}.

Traditionally in transfer learning for supervised and semi-supervised learning the \textit{domain} is defined as~$\mathcal{D} = \{\mathcal{X}, p(X)\}$, where $p(X)$ is the marginal distribution over the input dataset $X$ sampled from the input space $\mathcal{X}$, and the \textit{learning task} is defined as~$\mathcal{T} = \{\mathcal{Y}, p(Y \vert X)\}$, where $p(Y \vert X)$ is the distribution of the outputs $Y$ from the output space $\mathcal{Y}$ given the inputs~\cite{pan2009survey}. 
In transfer learning, there is a \textit{source} task $\mathcal{T}_s$ and domain $\mathcal{D}_s$ for which a model is originally optimized, and a \textit{target} task $\mathcal{T}_t$ and domain $\mathcal{D}_t$ on which that model's performance will be measured. 

This high-level designation of source and target remains for transfer in RL as well. 
However, as the environment is a decision process to be sampled through interaction instead of a pre-sampled set, the formulation of $\mathcal{D}$ and $\mathcal{T}$ is distributed according to the sampling procedure, which is non-i.i.d as it is a function of the policy and, by association,  exploration method.
Therefore, when discussing transfer in model-free RL like Q-learning, the learning domain becomes $\mathcal{D} = \{\mathcal{S}, P\left(s^{\prime}, r \mid s, a\right) \}$ and the learning task becomes~$\mathcal{T} = \{\mathcal{A}, Q(S_t, A_t)\}$ 

\begin{table*}[ht]
\centering
\begin{tabular}{c|p{3cm}|p{3cm}|p{3cm}|}
\multirow{2}{*}{\bf Algorithmic Instantiation} & \multicolumn{3}{c|}{\bf Exploration Principle}\\
\cline{2-4}
 & \textbf{Randomness} & \textbf{Prediction Error} & \textbf{Balanced Sampling} \\ 
\cline{1-4}
\textbf{Intrinsic Reward}  & \cite{DBLP:journals/corr/abs-1810-12894} [RND] & \cite{Oudeyer07article} [IAC], \cite{10.5555/3305890.3305968}) [ICM] & \cite{10.5555/3157096.3157262} [CTS-DQN]    \\ 
\hline
\textbf{Loss Function}  & \cite{ziebart2010modeling}[MCE], \cite{haarnoja2018soft}[SAC], \cite{hazan2019provably}[MEE] & \cite{dvzeroski2001relational}[RRL], \cite{zambaldi2018deep}[RIB] & BS \cite{azizzadenesheli2018efficient}[BDQN]   \\ 
\hline
\textbf{Multi-Policy/ Multi-Goal}  & \cite{sutton2018reinforcement}$\varepsilon$-greedy, \cite{nair2018visual}[RIG]  
& \cite{NIPS2017_453fadbd}[HER], \cite{NEURIPS2019_532b7cbe}[VHER] & \cite{wiering1998efficient}[UCB], \cite{MAL-070}[TS]  
\\ 
\hline
\end{tabular}
\caption{This table lays out our taxonomy of exploration for transfer, with algorithms that exemplify each exploration principle--algorithmic integration combination. } 
\label{taxonomy}
\end{table*}

\section{Exploration for Online Task Transfer}

This paper concerns itself with what we call the \textit{online task transfer in reinforcement learning} problem (OTTRL).
Given a model optimized for an original source learning task $\mathcal{T}_s$ and domain $D_s$ without any information about the target task except that a target task exists, we would like to optimize the model's performance on the target learning task $\mathcal{T}_t$ in an unknown target domain $D_t$ with as few training iterations and samples as possible.
This is distinct from other important problems in transfer learning such as \textit{model transfer learning} (sometimes referred to as knowledge distillation~\cite{hinton2015distilling,gou2021knowledge} or teacher-student frameworks~\cite{torrey2013teaching,zhan2015online} ), where target task knowledge encoded in a ``teacher'' agent is transferred by the teacher into a ``student'' agent with the same target task and domain.
OTTRL is also distinct in that, unlike most cases in domain adaptation and skill transfer~\cite{taylor2009transfer}, the target task and domain are not known in advance.

To better understand existing exploration techniques in the context of OTTRL, we propose a taxonomy of exploration for transfer.
We characterize exploration methods by two axes: 
\begin{itemize}
    \item \textbf{Algorithmic instantiation} 
    characterizes the mechanism within the RL algorithm that determines how the exploration-exploitation tradeoff is achieved;
    \item \textbf{Exploration principle}
    characterizes an agent's behavior beyond greedily selecting actions to maximize reward.
\end{itemize}

%
We consider three means of algorithmic instantiation: \textit{intrinsic reward}, a modified \textit{loss function}, and using \textit{multiple goals or policies} within the single learning task. 
Intrinsic motivation is a quality of exploration methods that incentivise visitation of sub-optimal transitions by reweighting the rewards experienced by the agent at those transitions. 
Modifying the reinforcement update or loss function is a tactic used by exploration methods that add exploration incentives fundamentally to the MDP or the optimization procedure. 
Lastly, methods that use multiple policies or imagined goals incentivises exploration by algorithmically shifting way actions are selected or goal-based rewards are generated.

We also consider three exploration principles: the integration of \textit{randomness} into the learning process, \textit{prediction error}, and \textit{balanced sampling} of the environment. 
There are many ways to use randomness in exploration, whether by injecting random noise into the input or some intermediate weight layer, using a stochastic policy, distilling a random network, or simply selecting random actions.
Prediction error methods identifying changes and novelty in the environment using a model, usually separate from the policy model, that explores more when it incorrectly predicts transition information.
Lastly, balanced sampling methods drive the agent toward uncertain or stale transitions, effectively trying to encourage models to experience all parts of the domain equally.

The following are exemplar exploration methods for each area of the taxonomy
(as seen in Table \ref{taxonomy}):

\begin{itemize}
    \item \textit{(Intrinsic Reward, Randomness)}: Random Network Distillation (RND)~\cite{DBLP:journals/corr/abs-1810-12894} sets an intrinsic reward equal to the error of a forward model's ability to match the projection of a random neural network. This is effectively using a random projection to build a density model: if the prediction error is low, that can be interpreted as something the network has seen before.
    \item \textit{(Intrinsic Reward, Prediction Error)}: Intrinsic Curiosity Module (ICM)~\cite{10.5555/3305890.3305968} learns a forward model and an inverse dynamics model (predicting the action from the previous and current states), and then adds an intrinsic reward based on the Euclidean distance between the state encoding by a forward model and the inverse model state encoding.
    \item \textit{(Intrinsic Reward, Balanced Sampling)}: Count-based deep models use density estimation methods such as Context Switching Trees~\cite{10.5555/3157096.3157262} to add intrinsic value to all states based on how frequently they have been visited. This biases greedy action selection to be more exploratory.   
    \item \textit{(Loss function, Randomness)}: Maximum entropy methods such as Maximum Entropy Exploration (MEE)~\cite{hazan2019provably} modify the loss function to have as high-entropy a policy as possible. The theory is that this will help learn a policy, with or without reward, which moves the policy to be more uniformly distributed over state space. 
    \item \textit{(Loss function, Prediction Error)}: Relational reasoning methods of exploration attempt to predict the connections between parts of the states, and methods like Relational reinforcement learning (RRL)~\cite{dvzeroski2001relational} reconfigure the loss so that it may work over MDPs described in a relational or logical language.
    \item \textit{(Loss function, Balanced Sampling)}: Reinforcement methods that use Bayesian approaches like Bayesian deep Q-networks (BDQN) modify the update and loss functions so that the agent can sample actions according to some posterior  uncertainty distribution.   
    \item \textit{(Multi-Policy, Randomness)}: The simplest exploration technique and the same one used by Q-learning and deep Q-learning illustrates this regime: $\varepsilon$-greedy. The agent has another policy, random selection, that is utilized at the frequency of the decaying hyperparameter $\varepsilon$.
    \item \textit{(Multi-Policy, Prediction Error)}: Instead of simply accepting failed trajectories as part of the reinforcement learning process, Hindsight experience replay (HER)~\cite{NIPS2017_453fadbd} trains goal-conditioned policies, where it considers the task goal to be the only one possible, then imagines and tries to predict the goals that would make failed trajectories successful.
    \item \textit{(Multi-Policy, Balanced Sampling)}: Thompson Sampling~\cite{MAL-070} (TS) and Upper Confidence Bound (UCB) switch back and forth between greedy action selection and a `policy' sampling procedure. Thomson sampling selects actions according to the likelihood that they are optimal, and UCB shares Bayesian regret bounds with Thompson sampling as well~\cite{NIPS2013_41bfd20a}. 
    As a result, these methods act to drive the policy to explore areas of uncertainty. 
\end{itemize}







\section{Discussion and Desiderata}

Based on the qualities of successful task transfer approaches~\cite{taylor2009transfer}, we propose that exploration methods designed for online task transfer should exhibit the following qualities:

\begin{enumerate}
    \item Exploration should always be engaged---even after the policy for the source task has converged---because online transfer means the agent must be ready to switch to the target task at any time.
    \item The exploration method should be designed to optimize for the source task performance, and then optimize for the target task only after the task switches; don't compromise or ignore source task performance to be good at the target task.
    \item Exploration methods should encourage generalization; they should help prevent the policy from overfitting to the source task. 
    \item Methods should generalize to as many tasks as possible and require little-to-no task-specific tuning.
\end{enumerate}

Applying these criteria to the taxonomy, we find that no single existing exploration technique is clearly best suited for OTTRL.
Among the algorithmic instantiations, intrinsic rewards often need to be tuned with respect to the extrinsic reward of the source task; to transfer successfully,  the target task reward would need to be similar in scale to the source task reward. 
Similarly, changes to the loss function may not generalize to the broad spectrum of target tasks; for example, with Bayesian methods, the prior for the source task may be well modeled by a normal Guassian, but the target task may not be. 
The multi-policy/multi-goal approach exhibits a strong ability to generalize, but may require tuning to not damage source task performance.

Among the exploration principles, randomness is naturally task agnostic, but many sources of randomness need to be designed to decay throughout training to ensure that the source policy converges. 
Balanced sampling on the other hand can remain active after convergence, but have downsides as they are usually anchored to the behavior of the policy. 
For example, assuming a problem where the source task is a maze and the target task is the same maze but with a shortcut added,
the optimal solution from the source task is still reachable for the policy, but is not optimal for the target task. 
Methods like prediction error that rely on the policy for selecting actions can be slow to adapt or possibly never converge to the target optimum. 
Moreover, prediction-based methods can be susceptible the ``noisy TV'' problem~\cite{DBLP:journals/corr/abs-1810-12894}. In a ``noisy TV'' environment there is a ``TV'' area that is continuously showing random novel images;  this scenario, an agent seeking ``newness'' through prediction error naively will just ``learn to watch TV.'' 



The opportunities and limitations identified through the taxonomy suggest that new exploration techniques may be required to improve online task transfer learning. 
A simple solution may be to combine these distinct approaches so that the strengths of one taxonomy area can compensate for the deficits of another.  
We find that the vast majority of exploration methods only exist in one area on this taxonomy.
While exploration methods from different parts of the taxonomy are often theoretically complementary, it can be non-trivial to integrate multiple methods in a way that increases general performance~\cite{hessel2018rainbow}.
Investigating the viability of mixed methods across the taxonomy is a suggested future avenue of research.

One subdomain of reinforcement learning research that has seen renewed recent attention and has potential to solve this ``mixing'' problem is the area of model-building control systems, also called \textit{world models}~\cite{schmidhuber1991curious, schmidhuber2015learning, ha2018recurrent, hafner2019dream}.
While model-based reinforcement learning techniques (e.g.,~\cite{silver2017mastering, pmlr-v97-hafner19a}) learn a forward model of the transition function and then use an algorithm like Monte Carlo Tree Search or Model Predictive Control to identify the best next action, world models learn models of the transition function and the policy together. 
In training of world models, the input to the policy is affected by the output of the forward model, and the policy directs the agent to provide learning samples for the forward model. 
As a result, world models are a natural way to combine the affect multiple exploration techniques on a single agent, where one technique can affect the policy and another the forward model. 
In their default configuration, the forward models still can fall into the trap of overfitting to the part of environment most sampled by the policy. 
However, as the optimization of forward model is only indirectly connected to the the optimization of the policy, world models have potential to overcome the challenges of other prediction-based exploration methods with regard to online transfer. 
There has been some recent work in methods of exploration in world models~\cite{sekar2020planning, mendonca2021discovering}, but we believe more investigation into their world model-focused exploration methods is warranted.

Another topic for consideration is mapping areas of the taxonomy to environmental or value-based considerations such as tolerance for risk during exploration for transfer. 
For example, if a robot controlled by a reinforcement learning agent is helping in a disaster zone, this robot may benefit from an exploration method that is \textit{(Multi-Policy/Multi-Goal, Prediction error)}-type. 
A technique from this area of the taxonomy will reduce unnecessary random actions by the robot, which could be dangerous, while preparing the robot to change its goal as needed. 
By matching the needs of an OTTRL application with an exploration method described by the taxonomy, research can accelerate the application of reinforcement learning to our open, ever-changing world.

\section{Other Related Work}

There exist many comprehensive and excellent reviews of exploration methods ~\cite{https://doi.org/10.48550/arxiv.2109.06668, amin2021survey}.
In this paper, we focused on prior work in the context of effective transfer learning.
There are prior works that do not fit neatly into this context and our taxonomy.
End-to-end world models~\cite{pmlr-v97-hafner19a,hafner2022deep} could be argued to offer both stochastic exploration native to policy learning and a loss change since loss from the world model propagates back to the policy by way of imagination-based training. 
Meta-learning of exploration methods~\cite{gupta2018exemplars} also does not fit perfectly.
While meta-learning may, for a given problem, employ one exploration technique characterized in our taxonomy, the exploration technique may be different for different domains.

Another means of ``transfer'' is training a model on multiple source learning tasks in advance of the target task so as to diversify the agent's experience~\cite{teh2017distral}.  
These multi-task learning methods are usually inferior to single task performance~\cite{vithayathil2020survey}, often using one-policy-per-task as an upper bound of performance~\cite{sodhani2021multi}. 
Additionally, they do not try to adapt to real-world perturbations instead trying to be robust to them. 

Work on novelty adaptation in RL has seen recent attention~\cite{langley2020open,boult2021towards,goel2021novelgridworlds,balloch2022novgrid}. 
Novelty is generally defined as a sudden but individual change---not necessarily directly observable---in the fundamental mechanics of the environment, leaving the high-level task the same. 
As such, novelty is a subdomain of task transfer that would likely benefit greatly from improved exploration methods.

\bibliographystyle{IEEEtran} 
\bibliography{references}

\begin{thebibliography}{10}
\providecommand{\url}[1]{#1}
\csname url@samestyle\endcsname
\providecommand{\newblock}{\relax}
\providecommand{\bibinfo}[2]{#2}
\providecommand{\BIBentrySTDinterwordspacing}{\spaceskip=0pt\relax}
\providecommand{\BIBentryALTinterwordstretchfactor}{4}
\providecommand{\BIBentryALTinterwordspacing}{\spaceskip=\fontdimen2\font plus
\BIBentryALTinterwordstretchfactor\fontdimen3\font minus
  \fontdimen4\font\relax}
\providecommand{\BIBforeignlanguage}[2]{{%
\expandafter\ifx\csname l@#1\endcsname\relax
\typeout{** WARNING: IEEEtran.bst: No hyphenation pattern has been}%
\typeout{** loaded for the language `#1'. Using the pattern for}%
\typeout{** the default language instead.}%
\else
\language=\csname l@#1\endcsname
\fi
#2}}
\providecommand{\BIBdecl}{\relax}
\BIBdecl

\bibitem{perkins1992transfer}
D.~N. Perkins, G.~Salomon \emph{et~al.}, ``Transfer of learning,''
  \emph{International encyclopedia of education}, vol.~2, pp. 6452--6457, 1992.

\bibitem{hopfield2010neurodynamics}
J.~J. Hopfield, ``Neurodynamics of mental exploration,'' \emph{Proceedings of
  the National Academy of Sciences}, vol. 107, no.~4, pp. 1648--1653, 2010.

\bibitem{mccloskey1989catastrophic}
M.~McCloskey and N.~J. Cohen, ``Catastrophic interference in connectionist
  networks: The sequential learning problem,'' in \emph{Psychology of learning
  and motivation}.\hskip 1em plus 0.5em minus 0.4em\relax Elsevier, 1989,
  vol.~24, pp. 109--165.

\bibitem{agarwal2022beyond}
R.~Agarwal, M.~Schwarzer, P.~S. Castro, A.~Courville, and M.~G. Bellemare,
  ``Beyond tabula rasa: Reincarnating reinforcement learning,'' in
  \emph{Advances in Neural Information Processing Systems}, 2022.

\bibitem{sutton2018reinforcement}
R.~S. Sutton and A.~G. Barto, \emph{Reinforcement learning: An
  introduction}.\hskip 1em plus 0.5em minus 0.4em\relax MIT press, 2018.

\bibitem{mnih2015human}
V.~Mnih, K.~Kavukcuoglu, D.~Silver, A.~A. Rusu, J.~Veness, M.~G. Bellemare,
  A.~Graves, M.~Riedmiller, A.~K. Fidjeland, G.~Ostrovski \emph{et~al.},
  ``Human-level control through deep reinforcement learning,'' \emph{nature},
  vol. 518, no. 7540, pp. 529--533, 2015.

\bibitem{taylor2009transfer}
M.~E. Taylor and P.~Stone, ``Transfer learning for reinforcement learning
  domains: A survey.'' \emph{Journal of Machine Learning Research}, vol.~10,
  no.~7, 2009.

\bibitem{Lazaric2012}
A.~Lazaric, \emph{Transfer in Reinforcement Learning: A Framework and a
  Survey}.\hskip 1em plus 0.5em minus 0.4em\relax Springer Berlin Heidelberg,
  2012, pp. 143--173.

\bibitem{zhu2020transfer}
\BIBentryALTinterwordspacing
Z.~Zhu, K.~Lin, A.~K. Jain, and J.~Zhou, ``Transfer learning in deep
  reinforcement learning: A survey,'' 2020. [Online]. Available:
  \url{https://arxiv.org/abs/2009.07888}
\BIBentrySTDinterwordspacing

\bibitem{chen2021cross}
X.-H. Chen, S.~Jiang, F.~Xu, Z.~Zhang, and Y.~Yu, ``Cross-modal domain
  adaptation for cost-efficient visual reinforcement learning,'' \emph{Advances
  in Neural Information Processing Systems}, vol.~34, pp. 12\,520--12\,532,
  2021.

\bibitem{pan2009survey}
S.~J. Pan and Q.~Yang, ``A survey on transfer learning,'' \emph{IEEE
  Transactions on knowledge and data engineering}, vol.~22, no.~10, pp.
  1345--1359, 2009.

\bibitem{DBLP:journals/corr/abs-1810-12894}
\BIBentryALTinterwordspacing
Y.~Burda, H.~Edwards, A.~J. Storkey, and O.~Klimov, ``Exploration by random
  network distillation,'' \emph{CoRR}, vol. abs/1810.12894, 2018. [Online].
  Available: \url{http://arxiv.org/abs/1810.12894}
\BIBentrySTDinterwordspacing

\bibitem{Oudeyer07article}
P.-Y. Oudeyer, F.~Kaplan, and V.~Hafner, ``Intrinsic motivation systems for
  autonomous mental development,'' \emph{Evolutionary Computation, IEEE
  Transactions on}, vol.~11, pp. 265 -- 286, 05 2007.

\bibitem{10.5555/3305890.3305968}
D.~Pathak, P.~Agrawal, A.~A. Efros, and T.~Darrell, ``Curiosity-driven
  exploration by self-supervised prediction,'' in \emph{Proceedings of the 34th
  International Conference on Machine Learning - Volume 70}, ser.
  ICML'17.\hskip 1em plus 0.5em minus 0.4em\relax JMLR.org, 2017, p.
  2778–2787.

\bibitem{10.5555/3157096.3157262}
M.~G. Bellemare, S.~Srinivasan, G.~Ostrovski, T.~Schaul, D.~Saxton, and
  R.~Munos, ``Unifying count-based exploration and intrinsic motivation,'' in
  \emph{Proceedings of the 30th International Conference on Neural Information
  Processing Systems}, 2016.

\bibitem{ziebart2010modeling}
B.~D. Ziebart, \emph{Modeling purposeful adaptive behavior with the principle
  of maximum causal entropy}.\hskip 1em plus 0.5em minus 0.4em\relax Carnegie
  Mellon University, 2010.

\bibitem{haarnoja2018soft}
T.~Haarnoja, A.~Zhou, P.~Abbeel, and S.~Levine, ``Soft actor-critic: Off-policy
  maximum entropy deep reinforcement learning with a stochastic actor,'' in
  \emph{International conference on machine learning}.\hskip 1em plus 0.5em
  minus 0.4em\relax PMLR, 2018, pp. 1861--1870.

\bibitem{hazan2019provably}
E.~Hazan, S.~Kakade, K.~Singh, and A.~Van~Soest, ``Provably efficient maximum
  entropy exploration,'' in \emph{International Conference on Machine
  Learning}.\hskip 1em plus 0.5em minus 0.4em\relax PMLR, 2019, pp. 2681--2691.

\bibitem{dvzeroski2001relational}
S.~D{\v{z}}eroski, L.~De~Raedt, and K.~Driessens, ``Relational reinforcement
  learning,'' \emph{Machine learning}, vol.~43, no.~1, pp. 7--52, 2001.

\bibitem{zambaldi2018deep}
V.~Zambaldi, D.~Raposo, A.~Santoro, V.~Bapst, Y.~Li, I.~Babuschkin, K.~Tuyls,
  D.~Reichert, T.~Lillicrap, E.~Lockhart, M.~Shanahan, V.~Langston, R.~Pascanu,
  M.~Botvinick, O.~Vinyals, and P.~Battaglia, ``Deep reinforcement learning
  with relational inductive biases,'' in \emph{International Conference on
  Learning Representations}, 2019.

\bibitem{azizzadenesheli2018efficient}
K.~Azizzadenesheli, E.~Brunskill, and A.~Anandkumar, ``Efficient exploration
  through bayesian deep q-networks,'' in \emph{2018 Information Theory and
  Applications Workshop (ITA)}.\hskip 1em plus 0.5em minus 0.4em\relax IEEE,
  2018, pp. 1--9.

\bibitem{nair2018visual}
A.~V. Nair, V.~Pong, M.~Dalal, S.~Bahl, S.~Lin, and S.~Levine, ``Visual
  reinforcement learning with imagined goals,'' \emph{Advances in neural
  information processing systems}, vol.~31, 2018.

\bibitem{NIPS2017_453fadbd}
M.~Andrychowicz, F.~Wolski, A.~Ray, J.~Schneider, R.~Fong, P.~Welinder,
  B.~McGrew, J.~Tobin, O.~Pieter~Abbeel, and W.~Zaremba, ``Hindsight experience
  replay,'' in \emph{Advances in Neural Information Processing Systems},
  I.~Guyon, U.~V. Luxburg, S.~Bengio, H.~Wallach, R.~Fergus, S.~Vishwanathan,
  and R.~Garnett, Eds., vol.~30.\hskip 1em plus 0.5em minus 0.4em\relax Curran
  Associates, Inc., 2017.

\bibitem{NEURIPS2019_532b7cbe}
H.~Sahni, T.~Buckley, P.~Abbeel, and I.~Kuzovkin, ``Addressing sample
  complexity in visual tasks using her and hallucinatory gans,'' in
  \emph{Advances in Neural Information Processing Systems}, H.~Wallach,
  H.~Larochelle, A.~Beygelzimer, F.~d\textquotesingle Alch\'{e}-Buc, E.~Fox,
  and R.~Garnett, Eds., vol.~32.\hskip 1em plus 0.5em minus 0.4em\relax Curran
  Associates, Inc., 2019.

\bibitem{wiering1998efficient}
M.~Wiering and J.~Schmidhuber, ``Efficient model-based exploration,'' in
  \emph{Proceedings of the Sixth International Conference on Simulation of
  Adaptive Behavior: From Animals to Animats}, vol.~6.\hskip 1em plus 0.5em
  minus 0.4em\relax MIT Press Cambridge, MA, 1998, pp. 223--228.

\bibitem{MAL-070}
\BIBentryALTinterwordspacing
D.~J. Russo, B.~V. Roy, A.~Kazerouni, I.~Osband, and Z.~Wen, ``A tutorial on
  thompson sampling,'' \emph{Foundations and Trends® in Machine Learning},
  vol.~11, no.~1, pp. 1--96, 2018. [Online]. Available:
  \url{http://dx.doi.org/10.1561/2200000070}
\BIBentrySTDinterwordspacing

\bibitem{hinton2015distilling}
G.~Hinton, O.~Vinyals, J.~Dean \emph{et~al.}, ``Distilling the knowledge in a
  neural network,'' \emph{arXiv preprint arXiv:1503.02531}, vol.~2, no.~7,
  2015.

\bibitem{gou2021knowledge}
J.~Gou, B.~Yu, S.~J. Maybank, and D.~Tao, ``Knowledge distillation: A survey,''
  \emph{International Journal of Computer Vision}, vol. 129, no.~6, pp.
  1789--1819, 2021.

\bibitem{torrey2013teaching}
L.~Torrey and M.~Taylor, ``Teaching on a budget: Agents advising agents in
  reinforcement learning,'' in \emph{Proceedings of the 2013 international
  conference on Autonomous agents and multi-agent systems}, 2013, pp.
  1053--1060.

\bibitem{zhan2015online}
Y.~Zhan and M.~E. Taylor, ``Online transfer learning in reinforcement learning
  domains,'' in \emph{2015 AAAI Fall Symposium Series}, 2015.

\bibitem{NIPS2013_41bfd20a}
D.~Russo and B.~Van~Roy, ``Eluder dimension and the sample complexity of
  optimistic exploration,'' in \emph{Advances in Neural Information Processing
  Systems}, C.~Burges, L.~Bottou, M.~Welling, Z.~Ghahramani, and K.~Weinberger,
  Eds., vol.~26.\hskip 1em plus 0.5em minus 0.4em\relax Curran Associates,
  Inc., 2013.

\bibitem{hessel2018rainbow}
M.~Hessel, J.~Modayil, H.~Van~Hasselt, T.~Schaul, G.~Ostrovski, W.~Dabney,
  D.~Horgan, B.~Piot, M.~Azar, and D.~Silver, ``Rainbow: Combining improvements
  in deep reinforcement learning,'' in \emph{Thirty-second AAAI conference on
  artificial intelligence}, 2018.

\bibitem{schmidhuber1991curious}
J.~Schmidhuber, ``Curious model-building control systems,'' in \emph{Proc.
  international joint conference on neural networks}, 1991, pp. 1458--1463.

\bibitem{schmidhuber2015learning}
------, ``On learning to think: Algorithmic information theory for novel
  combinations of reinforcement learning controllers and recurrent neural world
  models,'' \emph{arXiv preprint arXiv:1511.09249}, 2015.

\bibitem{ha2018recurrent}
D.~Ha and J.~Schmidhuber, ``Recurrent world models facilitate policy
  evolution,'' \emph{Advances in neural information processing systems},
  vol.~31, 2018.

\bibitem{hafner2019dream}
D.~Hafner, T.~Lillicrap, J.~Ba, and M.~Norouzi, ``Dream to control: Learning
  behaviors by latent imagination,'' in \emph{International Conference on
  Learning Representations}, 2019.

\bibitem{silver2017mastering}
D.~Silver, J.~Schrittwieser, K.~Simonyan, I.~Antonoglou, A.~Huang, A.~Guez,
  T.~Hubert, L.~Baker, M.~Lai, A.~Bolton \emph{et~al.}, ``Mastering the game of
  go without human knowledge,'' \emph{nature}, vol. 550, no. 7676, pp.
  354--359, 2017.

\bibitem{pmlr-v97-hafner19a}
D.~Hafner, T.~Lillicrap, I.~Fischer, R.~Villegas, D.~Ha, H.~Lee, and
  J.~Davidson, ``Learning latent dynamics for planning from pixels,'' in
  \emph{Proceedings of the 36th International Conference on Machine Learning},
  K.~Chaudhuri and R.~Salakhutdinov, Eds., vol.~97, 2019, pp. 2555--2565.

\bibitem{sekar2020planning}
R.~Sekar, O.~Rybkin, K.~Daniilidis, P.~Abbeel, D.~Hafner, and D.~Pathak,
  ``Planning to explore via self-supervised world models,'' in
  \emph{International Conference on Machine Learning}.\hskip 1em plus 0.5em
  minus 0.4em\relax PMLR, 2020, pp. 8583--8592.

\bibitem{mendonca2021discovering}
R.~Mendonca, O.~Rybkin, K.~Daniilidis, D.~Hafner, and D.~Pathak, ``Discovering
  and achieving goals via world models,'' \emph{Advances in Neural Information
  Processing Systems}, vol.~34, pp. 24\,379--24\,391, 2021.

\bibitem{https://doi.org/10.48550/arxiv.2109.06668}
\BIBentryALTinterwordspacing
T.~Yang, H.~Tang, C.~Bai, J.~Liu, J.~Hao, Z.~Meng, P.~Liu, and Z.~Wang,
  ``Exploration in deep reinforcement learning: A comprehensive survey,'' 2021.
  [Online]. Available: \url{https://arxiv.org/abs/2109.06668}
\BIBentrySTDinterwordspacing

\bibitem{amin2021survey}
S.~Amin, M.~Gomrokchi, H.~Satija, H.~van Hoof, and D.~Precup, ``A survey of
  exploration methods in reinforcement learning,'' \emph{arXiv preprint
  arXiv:2109.00157}, 2021.

\bibitem{hafner2022deep}
D.~Hafner, K.-H. Lee, I.~Fischer, and P.~Abbeel, ``Deep hierarchical planning
  from pixels,'' \emph{arXiv preprint arXiv:2206.04114}, 2022.

\bibitem{gupta2018exemplars}
A.~Gupta, R.~Mendonca, Y.~Liu, P.~Abbeel, and S.~Levine, ``Meta-reinforcement
  learning of structured exploration strategies,'' in \emph{Advances in Neural
  Information Processing Systems}, S.~Bengio, H.~Wallach, H.~Larochelle,
  K.~Grauman, N.~Cesa-Bianchi, and R.~Garnett, Eds., vol.~31.\hskip 1em plus
  0.5em minus 0.4em\relax Curran Associates, Inc., 2018.

\bibitem{teh2017distral}
Y.~Teh, V.~Bapst, W.~M. Czarnecki, J.~Quan, J.~Kirkpatrick, R.~Hadsell,
  N.~Heess, and R.~Pascanu, ``Distral: Robust multitask reinforcement
  learning,'' \emph{Advances in neural information processing systems},
  vol.~30, 2017.

\bibitem{vithayathil2020survey}
N.~Vithayathil~Varghese and Q.~H. Mahmoud, ``A survey of multi-task deep
  reinforcement learning,'' \emph{Electronics}, vol.~9, no.~9, p. 1363, 2020.

\bibitem{sodhani2021multi}
S.~Sodhani, A.~Zhang, and J.~Pineau, ``Multi-task reinforcement learning with
  context-based representations,'' in \emph{International Conference on Machine
  Learning}.\hskip 1em plus 0.5em minus 0.4em\relax PMLR, 2021, pp. 9767--9779.

\bibitem{langley2020open}
P.~Langley, ``Open-world learning for radically autonomous agents,'' in
  \emph{Proceedings of the AAAI Conference on Artificial Intelligence},
  vol.~34, 2020, pp. 13\,539--13\,543.

\bibitem{boult2021towards}
T.~Boult, P.~Grabowicz, D.~Prijatelj, R.~Stern, L.~Holder, J.~Alspector, M.~M.
  Jafarzadeh, T.~Ahmad, A.~Dhamija, C.~Li \emph{et~al.}, ``Towards a unifying
  framework for formal theories of novelty,'' in \emph{Proceedings of the AAAI
  Conference on Artificial Intelligence}, vol.~35, 2021, pp. 15\,047--15\,052.

\bibitem{goel2021novelgridworlds}
S.~Goel, G.~Tatiya, M.~Scheutz, and J.~Sinapov, ``Novelgridworlds: A benchmark
  environment for detecting and adapting to novelties in open worlds,''
  \emph{Adaptive and Learning Agents Workshop at AAMAS}, 2021.

\bibitem{balloch2022novgrid}
J.~Balloch, Z.~Lin, M.~Hussain, A.~Srinivas, X.~Peng, J.~Kim, and M.~Riedl,
  ``Novgrid: A flexible grid world for evaluating agent response to novelty,''
  in \emph{In Proceedings of AAAI Symposium, Designing Artificial Intelligence
  for Open Worlds}, 2022.

\end{thebibliography}

\end{document}